\documentclass[11pt,a4paper]{article}
\usepackage[hyperref]{acl2021}
\usepackage{times}
\usepackage{latexsym}
\usepackage{tabularx}
\usepackage{graphicx}
\usepackage{amsmath}
\usepackage{booktabs}
\usepackage{multirow}
\usepackage{subfig}
\usepackage{caption}
\usepackage{verbatim}

\usepackage{microtype}

\aclfinalcopy 
\def\aclpaperid{2443} 

\title{MMGCN: Multimodal Fusion via Deep Graph Convolution Network for Emotion Recognition in Conversation}
\author{Jingwen Hu, Yuchen Liu, Jinming Zhao, Qin Jin\footnotemark[1] \\
School of Information, Renmin University of China\\
{\tt\small \{hujingwen\_benja, Liuyuchen\_Alfred, zhaojinming, qjin\}@ruc.edu.cn}
}

\begin{document}
\maketitle
\renewcommand{\thefootnote}{\fnsymbol{footnote}}
\footnotetext[1]{Corresponding Author}
\renewcommand{\thefootnote}{\arabic{footnote}}

\begin{abstract}
Emotion recognition in conversation (ERC) is a crucial component in affective dialogue systems, which helps the system understand users' emotions and generate empathetic responses. 
However, most works focus on modeling speaker and contextual information primarily on the textual modality or simply leveraging multimodal information through feature concatenation. In order to explore a more effective way of utilizing both multimodal and long-distance contextual information, we propose a new model based on multimodal fused graph convolutional network, MMGCN, in this work. 
MMGCN can not only make use of multimodal dependencies effectively, but also leverage speaker information to model inter-speaker and intra-speaker dependency. We evaluate our proposed model on two public benchmark datasets, IEMOCAP and MELD, and the results prove the effectiveness of MMGCN, which outperforms other SOTA methods by a significant margin under the multimodal conversation setting.
\end{abstract}

\section{Introduction}
Emotion is an important part of human daily communication. Emotion Recognition in Conversation (ERC) aims to automatically identify and track the emotional status of speakers during a dialogue. It has attracted increasing attention from researchers in the field of natural language processing and multimodal processing. ERC has a wide range of potential applications such as assisting conversation analysis for legal trials and e-health services etc. It is also a key component for building natural human-computer interactions that can produce emotional responses in a dialogue.

\begin{figure}[t]
\centering
\includegraphics[scale=0.105]{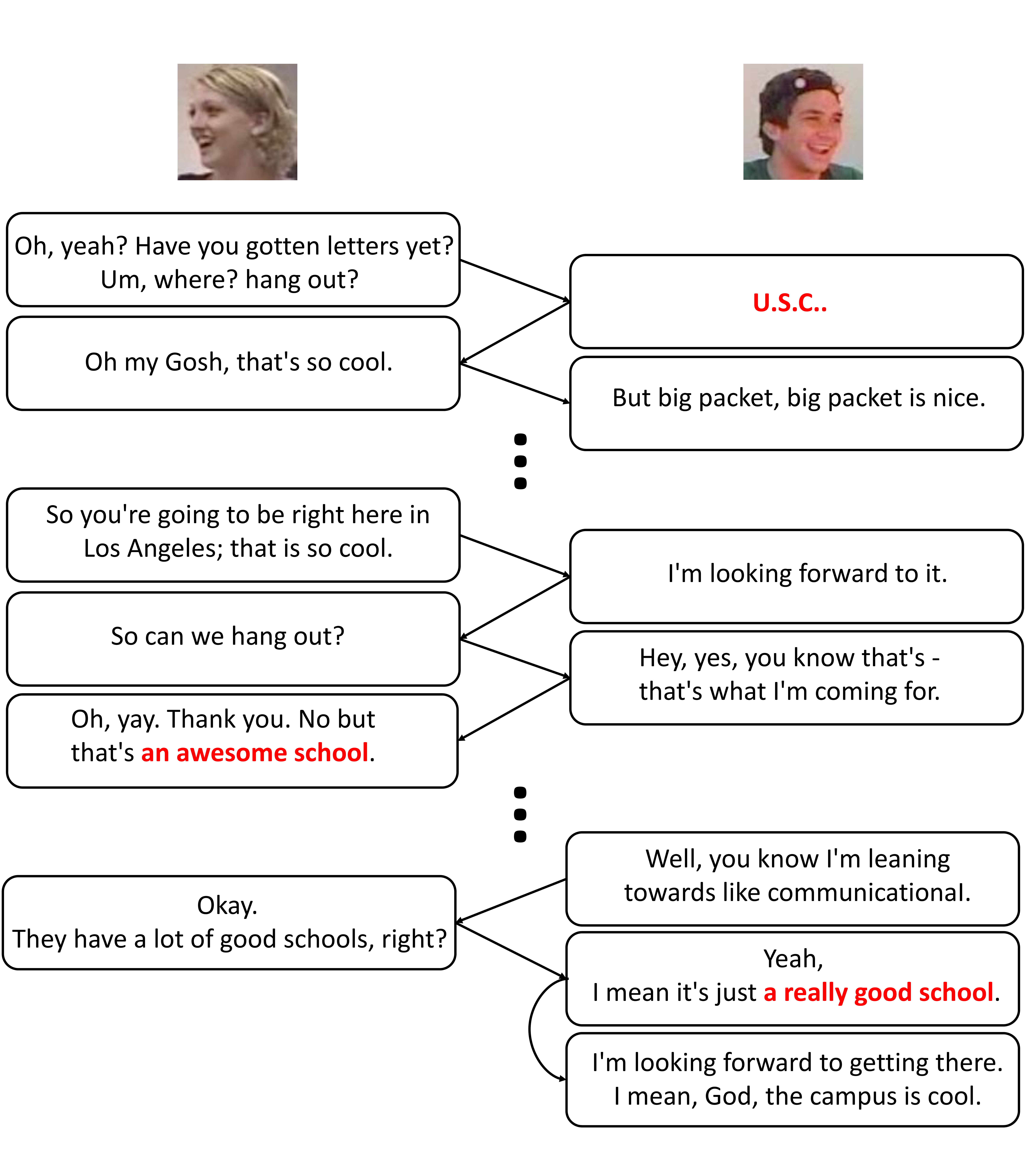}
\caption{Illustration of an example conversation in the IEMOCAP dataset}
\label{Fig:img1}
\end{figure}

The fast growing availability of conversational data on social media is one of the factors that boost the research focus on emotion recognition in conversation. Different from traditional emotion recognition on isolated utterances, emotion recognition in conversation requires context modeling of individual utterances. The context can be attributed to the preceding utterances, temporality in conversation turns, or speaker related information etc. Different models have been proposed to capture the contextual information in previous works, including the LSTM-based model \cite{poria2017context}, the conversational memory network (CMN) model \cite{hazarika2018conversational}, interactive conversational memory network (ICON) model \cite{hazarika2018icon}, and DialogueRNN model \cite{majumder2019dialoguernn} etc. 
In the example conversation as shown in Figure~\ref{Fig:img1}, the two speakers are chatting in the context of the male speaker being admitted to USC. In this chatting scene, they change topics a few times, such as the female speaker inviting the male speaker out to play and so on. But they keep coming back to the topic of USC, and then both of them express an excitement emotional status. It shows that long-distance contextual information is of great help to the prediction of speakers' emotions. 
However, previous models can not effectively capture both speaker and long-distance dialogue contextual information simultaneously in multi-speaker conversation scenarios. 
Ghosal et al.\cite{ghosal2019dialoguegcn}, therefore, first propose the DialogueGCN model which applies graph convolutional network (GCN) to capture long-distance contextual information in a conversation. DialogueGCN takes each utterance as a node and connects any nodes that are in the same window within a conversation. It can well model both the dialogue context and speaker information which leads to the state-of-the-art ERC performance. However, like most previous models, DialogGCN only focuses on the textual modality of the conversation, ignoring effective combination of other modalities such as visual and acoustic modalities. 
Works that consider multimodal contextual information often conduct the simple feature concatenation type of multimodal fusion. 

In order to effectively explore the multimodal information and at the same time capture long-distance contextual information, we propose a new multimodal fused graph convolutional network (MMGCN) model in this work. MMGCN constructs the fully connected graph in each modality, and builds edge connections between nodes corresponding to the same utterance across different modalities, so that contextual information across different modalities can interact. In addition, the speaker information is injected into MMGCN via speaker embedding. 
Furthermore, different from DialogueGCN, which is a non-spectral domain GCN and its many optimized matrices occupy too much computing resource, we encode the multimodal graph using spectral domain GCN and extend the GCN from a single layer to deep layers. To verify the effectiveness of the proposed model, we carry out experiments on two benchmark multimodal conversation datasets, IEMOCAP and MELD. MMGCN significantly outperforms other models on both datasets.

The rest of the paper is organized as follows: Section 2 discusses some related works; Section 3 introduces the proposed MMGCN model in details; Section 4 and 5 present the experiment setups on two public benchmark datasets and the analysis of experiment results and ablation study; Finally, Section 6 draws some conclusions.
\section{Related Work}
\subsection{Emotion Recognition in Conversation}

With the fast development of social media, much more interaction data become available, including several open-sourced conversation datasets such as IEMOCAP\cite{busso2008iemocap}, AVEC\cite{schuller2012avec}, MELD\cite{poria2018meld}, etc. ERC has attracted much research attention recently.

\begin{figure*}[t]
\centering
\includegraphics[scale=0.32]{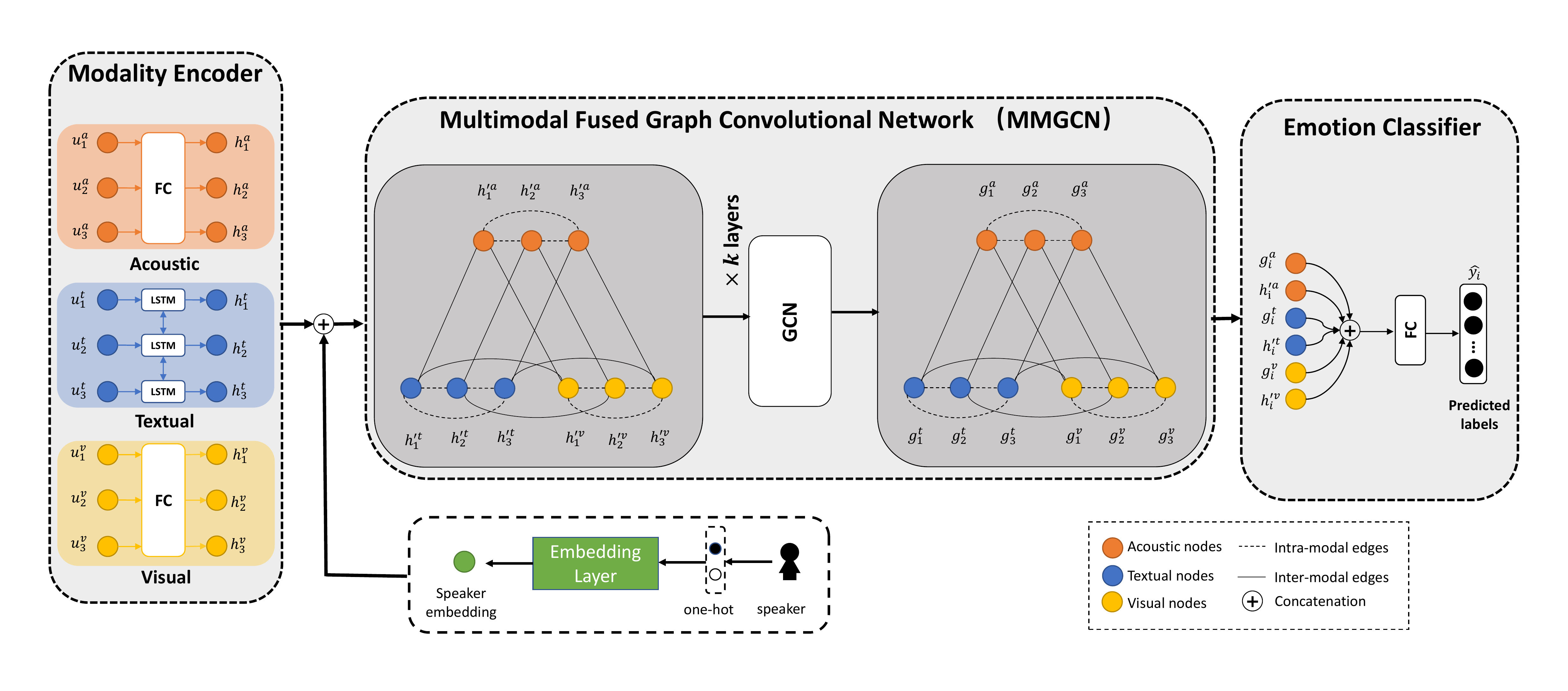}
\caption{Framework illustration of the MMGCN based emotion recognition in conversation, which consists of three key components: Modality Encoder, Multimodal Graph Convolutional Network, Emotion Classifier. }
\label{Fig:model}
\end{figure*}

Many previous works focus on modeling contextual information due to its importance in ERC. Poria et al.~\cite{poria2017context} leverage a LSTM-based model to capture interaction history context. Hazarika et al.~\cite{hazarika2018conversational,hazarika2018icon} first pay attention to the importance of speaker information and exploit different memory networks to model different speakers. DialogueRNN~\cite{majumder2019dialoguernn} leverage distinct GRUs to capture speakers' contextual information. DialogueGCN~\cite{ghosal2019dialoguegcn} construct the graph considering both speaker and conversation sequential information and achieve the state-of-the-art performance. 

\subsection{Multimodal Fusion}

Most recent studies on ERC focus primarily on the textual modality. \cite{poria2017context,hazarika2018conversational,hazarika2018icon} leverage multimodal information through concatenating features from three modalities without modeling the interaction between modalities. \cite{chen2017multimodal} conduct multimodal fusion at the word-level for emotion recognition of isolated utterances. \cite{sahay2018multimodal} consider contextual information and use relations in the emotion labels across utterances to predict the emotion
\cite{zadeh2018memory} propose MFN to fuse information of multi-views, which aligns features from different modalities well. However, MFN neglects to model speaker information, which is significant to ERC as well.
The state-of-the-art dialogueGCN model only considers the textual modality. In order to explore a more effective way of fusing multiple modalities and at the same time capturing contextual conversation information, we propose MMGCN which constructs a graph based on all three muoldalities. 

\subsection{Graph Convolutional Network}

Graph convolutional networks have been widely used in the past few years for their ability to cope with non-Euclidean data. Mainstream GCN methods can be divided into spectral domain methods and non-spectral domain methods~\cite{velivckovic2017graph}. Spectral domain GCN methods~\cite{zhang2019modeling} are based on Laplace Spectral decomposition theory. They can only deal with undirected graphs. Non-spectral domain GCN methods~\cite{velivckovic2017graph,schlichtkrull2018modeling,li2015gated} can be applied to both directed and undirected graphs, but consuming larger computing resource. Recently, researchers have proposed methods to make spectral domain GCN deeper without over-smoothing~\cite{li2019deepgcns,chen2020simple}. In order to further improve MMGCN on ERC, we encode the multimodal graph using spectral domain GCN with deep layers.

\section{Method}

A dialogue can be defined as a sequence of utterances $\{u_{1},u_{2},...,u_{N}\}$, where $N$ is the number of utterances. Each utterance involves three sources of utterance-aligned data corresponding to three modalities, including acoustic (a), visual (v) and textual (t) modalities, which can be represented as follows:

\begin{small}
\begin{equation}
    \centering
    u_{i} = \{u_{i}^a, u_{i}^v, u_{i}^t\} 
\end{equation}
\end{small}
where $u_{i}^a$, $u_{i}^v$, $u_{i}^t$ denote the raw feature representation of $u_{i}$ from the acoustic, visual and textual modality, respectively. The emotion recognition in conversation task aims to predict the emotional status label for each utterance $u_i$ in the conversation based on the available information from all three modalities. 
Figure~\ref{Fig:model} illustrates the overall framework of our proposed emotion recognition in conversation system, 
which consists of three key modules: Modality Encoder, Multimodal Fused Graph Convolutional Network (MMGCN), and Emotion Classifier. 



\subsection{Modality Encoder}
\label{sec:encoder}
As we mentioned above, the dialog context information is important for predicting the emotion label of each utterance. Therefore, it is beneficial to encode the contextual information into the utterance feature representation. We generate the context-aware utterance feature encoding for each modality through the corresponding modality encoder. To be specific, we apply a bidirectional Long Short Term Memory (LSTM) network to encode the sequential textual context information for the textual modality. For the acoustic and visual modalities, we apply a fully connected network. The context-aware feature encoding for each utterance can be formulated as follows:

\begin{small}
\begin{equation}
\begin{aligned}
    \centering
    &h_{i}^t = [\overrightarrow{\mathrm{LSTM}}(u_{i}^t,h_{i-1}^t),\overleftarrow{\mathrm{LSTM}}(u_{i}^t,h_{i+1}^t)] \\
    &h_{i}^a = W_{e}^a u_{i}^a+b_{i}^a \\
    &h_{i}^v = W_{e}^v u_{i}^v+b_{i}^v
\end{aligned}
\end{equation}
\end{small}
where $u_{i}^a$, $u_{i}^v$ , $u_{i}^t$ are the context-independent raw feature representation of utterance $i$ from the acoustic, visual and textual modalities, respectively. The modality encoder outputs the context-aware raw feature encoding $h_{i}^a$, $h_{i}^v$, and $h_{i}^t$ accordingly.
\subsection{Multimodal fused GCN (MMGCN)}

In order to capture the utterance-level contextual dependencies across multiple modalities, we propose a Multimodal fused Graph Convolutional Network (MMGCN). We construct a spectral domain graph convolutional network to encode the multimodal contextual information inspired by \cite{li2019deepgcns,chen2020simple}. We also stack more layers to construct a deep GCN. Furthermore, we add learned speaker-embeddings to encode the speaker-level contextual information.

\subsubsection{Speaker Embedding} 

As mentioned above, speaker information is important for ERC. In order to encode the speaker identity information, we add speaker embeddings to the features before constructing the graph. Assuming there are $M$ parties in a dialogue, then the size of the speaker embedding is $M$. We show a two-speaker conversation case in Figure ~\ref{Fig:model}. The original speaker identity can be denoted with a one-hot vector $s_{i}$ and the speaker embedding $S_{i}$ is calculated as follows:

\begin{small}
\begin{equation}
    \centering
        S_{i} = W_{s} s_{i} +b_{i}^s
\end{equation}
\end{small}
The speaker embedding can then be leveraged to attach speaker information in the graph construction.

\subsubsection{Graph Construction}  
\label{sec:graph}

A dialogue with $N$ utterances can be represented as an undirected graph $\mathcal{G}=(\mathcal{V}, \mathcal{E})$, where $\mathcal{V}$ ($\lvert \mathcal{V}\rvert = 3N$) denotes utterance nodes in three modalities and $\mathcal{E} \subset \mathcal{V} \times \mathcal{V}$ is a set of relationships containing context, speaker and modality dependency. We construct the graph as follows:

\noindent\textbf{Nodes:}
\noindent Each utterance is represented by three nodes $v_{i}^a$, $v_{i}^v$, $v_{i}^t$ in a graph, initialized with $h_{i}^{'a}$,$h_{i}^{'v}$,$h_{i}^{'l}$, which represent $[h_{i}^a, S_{i}]$, $[h_{i}^v, S_{i}]$, $[h_{i}^t, S_{i}]$ respectively, corresponding to the three modalities. Thus, given a dialogue with $N$ utterances, we construct a graph with $3N$ nodes.

\noindent\textbf{Edges:}
\noindent We assume that each utterance has certain connection to other utterances in the same dialogue. 
Therefore, any two nodes in the same modality in the same dialogue are connected in the graph. Furthermore, each node is connected with the nodes which correspond to the same utterance but from different modalities. For example, $v_{i}^a$ will be connected with $v_{i}^v$ and $v_{i}^t$ in the graph.

\noindent\textbf{Edge Weighting:} We assume that if two nodes have higher similarity, the information interaction between them is also more important, and the edge weight between them should be higher. In order to capture the similarities between node representations, following \cite{skianis2018fusing}, we use the angular similarity to represent the edge weight between two nodes.

There are two types of edges in the graph: 1) edges connecting nodes from the same modality, and 2) edges connecting nodes from different modalities. To differentiate them, we use different edge weighting strategies. For the first type of edges, the edge weight is computed as: 

\begin{small}
\begin{equation}
    \centering
        \mathcal{A}_{ij} = 1-\frac{arccos(sim(n_i, n_j))}{\pi}
\end{equation}
\end{small}
where $n_{i}$ and $n_{j}$ denote the feature representations of the $i$-th and $j$-th node in the graph. For the second type of edges, the edge weight is computed as:

\begin{small}
\begin{equation}
    \centering
        \mathcal{A}_{ij} = \gamma(1-\frac{arccos(sim(n_i, n_j))}{\pi})
\end{equation}
\end{small}
where $\gamma$ is a hyper parameter.

\noindent\textbf{Graph Learning:}
Inspired by \cite{chen2020simple}, we build a deep graph convolutional network based on the undirected graph formed following the above construction steps to further encode the contextual dependencies. To be specific, given the undirected graph $\mathcal{G}=(\mathcal{V}, \mathcal{E})$, let $\tilde{\mathcal{P}}$ be the renormalized graph Laplacian matrix \cite{kipf2016semi} of $\mathcal{G}$:

\begin{small}
\begin{equation}
    \centering
    \begin{aligned}
        \tilde{\mathcal{P}} &= \tilde{\mathcal{D}}^{-1/2}\tilde{\mathcal{A}}\tilde{\mathcal{D}}^{-1/2} \\
                  &= (\mathcal{D} + \mathcal{I})^{-1/2}(\mathcal{A} + \mathcal{I})(\mathcal{D} + \mathcal{I})^{-1/2} 
    \end{aligned}
\end{equation}
\end{small}
where $\mathcal{A}$ denotes the adjacency matrix, $\mathcal{D}$ denotes the diagonal degree matrix of graph $G$, and $\mathcal{I}$ denotes identity matrix. The iteration of GCN from different layers can be formulated as:

\begin{scriptsize}
\begin{equation}
    \centering
        \mathcal{H}^{(l+1)} =\sigma(((1-\alpha)\tilde{\mathcal{P}}\mathcal{H}^{(l)}+\alpha \mathcal{H}^{(0)})((1-\beta^{(l)})\mathcal{I}+\beta^{(l)} \mathcal{W}^{(l)})) 
\end{equation}
\end{scriptsize}
where $\alpha$ and $\beta^{(l)}$ are two hyper parameters, $\sigma$ denotes the activation function and $\mathcal{W}^{(l)}$ is a learnable weight matrix. To ensure the decay of the weight matrix adaptively increases when stacking more layers, we  set $\beta^{(l)} = \log(\frac{\eta}{l}+1)$, where $\eta$ is also a hyper parameter. A residual connection to the first layer $\mathcal{H}^{(0)}$ is added to the representation $\tilde{\mathcal{P}}\mathcal{H}^{(l)}$ and an identity mapping $\mathcal{I}$ is added to the weight matrix $\mathcal{W}^{(l)}$. With such residual connection, we can make MMGCN deeper to further improve performance.

\subsection{Emotion Classifier}
As described in sec.~\ref{sec:graph}, we initialize nodes with the combination of utterance feature and speaker embedding, $h^{'}_{i}$.

\begin{small}
\begin{equation}
    \centering
    h^{'}_{i} = [h^{'a}_{i}, h^{'v}_{i}, h^{'t}_{i}].
\end{equation}
\end{small}
Let $g_{i}^a$, $g_{i}^v$ and $g_{i}^t$ be the features of different modalities encoded by the GCN. The features corresponding to the same utterance are concatenated:

\begin{small}
\begin{equation}
    \centering
    g_{i} = [g_{i}^a, g_{i}^v, g_{i}^t].
\end{equation}
\end{small}
We then can concatenate $g_{i}$ and $h_{i}$ to generate the final feature representation for each utterance:

\begin{small}
\begin{align}
    \centering
    &e_{i} = [h^{'}_{i}, g_{i}] ,
\end{align}
\end{small}
$e_{i}$ is then fed into a MLP with fully connected layers to predict the emotion label $\hat{y_{i}}$ for the utterance:

\begin{small}
\begin{equation}
\begin{aligned}
    &l_{i} = RELU(W_{l}e_{i}+b_{l})\\
    &P_{i} = Softmax(W_{smax}l_{i}+b_{smax})\\
    &\hat{y_{i}} = \mathop{\arg\min}_{k} (P_{i}[k])
\end{aligned}
\end{equation}
\end{small}

\subsection{Training Objectives}

We use categorical cross-entropy along with L2-regularization as the loss function during training:

\begin{small}
\begin{equation}
    \centering
    \mathcal{L} = -\frac{1}{\sum_{s=1}^N c(s)} \sum_{i=1}^N \sum_{j=1}^{c(i)} log P_{i, j}[y_{i, j}] + \lambda \left\|\theta\right\|_{2}
\end{equation}
\end{small}
where $N$ is the number of dialogues, $c(i)$ is the number of utterances in dialogue $i$, $P_{i, j}$ is the probability distribution of predicted emotion labels of utterance $j$ in dialogue $i$, $y_{i, j}$ is the expected class label of utterance $j$ in dialogue $i$, $\lambda$ is the L2-regularization weight, and $\theta$ is the set of all trainable parameters.
We use stochastic gradient descent based Adam \cite{kingma2014adam} optimizer to train our network. 
Hyper parameters are optimized using grid search.

\section{Experiment Setups}
\subsection{Dataset}
We evaluate our proposed MMGCN model on two benchmark datasets, IEMOCAP\cite{busso2008iemocap} and MELD\cite{poria2018meld}. Both are multimodal datasets with aligned acoustical, visual and textual information of each utterance in a conversation. Followed \cite{ghosal2019dialoguegcn}, we partition both datasets into train and test sets with roughly 8:2 ratio. Table~\ref{tab:dataset} shows the distribution of train and test samples for both datasets. 

\textbf{IEMOCAP:} 
The dataset contains 12 hours of videos of two-way conversations from ten unique speakers, where only the first eight speakers from session one to four are used in the training set. Each video contains a single dyadic dialogue, segmented into utterances. There are in total 7433 utterances and 151 dialogues. Each utterance in the dialogue is annotated with an emotion label from six classes, including happy, sad, neutral, angry, excited and frustrated.

\textbf{MELD:} Multi-modal Emotion Lines Dataset (MELD)  is a multi-modal and multi-speaker conversation dataset. Compared to the Emotion Lines dataset \cite{chen2018emotionlines}, MELD has three modality-aligned conversation data with higher quality. There are in total 13708 utterances, 1433 conversations and 304 different speakers. Specifically, different from dyadic conversation datasets such as IEMOCAP, MELD has three or more speakers in a conversation. Each utterance in the dialogue is annotated with an emotion label from seven classes, including anger, disgust, fear, joy, neutral, sadness and surprise.

\begin{center}
    \begin{table}[t]
        \centering
        \scalebox{0.9}{
        \begin{tabular}{c|cc|cc}
            \hline
            \multirow{2}*{Dataset}& \multicolumn{2}{c|}{dialogues} & \multicolumn{2}{c}{utterances} \\
            ~ & {train+val} & {test} & {train+val} & {test}\\ \hline
            IEMOCAP           &{120}     &31    &{5810}     &1623  \\ 
            MELD           &1153      &280     &11098       &2610  \\ \hline
        \end{tabular}}
        \caption{Data distribution of IEMOCAP and MELD}
        \label{tab:dataset}
    \end{table}
\end{center}

\begin{table*}[]
    \centering
    \scalebox{0.9}{
    \begin{tabular}{c|cccccc|c||c}
        \hline
        \multirow{2}*{}& \multicolumn{7}{c||}{IEMOCAP} & \multicolumn{1}{c}{MELD}\\
        \cline{2-9}
        ~ & {Happy}& {Sad}& {Neutral}& {Angry}&{Excited}&{Frustrated}&{\textbf{Average(w)}}&{\textbf{Average(w)}}\\
        \hline
        BC-LSTM       &34.43     &60.87        &51.81        &56.73       &57.95       &58.92       &54.95          &56.80 \\ 
        CMN           &30.38     &62.41        &52.39        &59.83       &60.25       &60.69       &56.13          &- \\
        ICON          &29.91     &64.57        &57.38        &63.04       &63.42       &60.81       &58.54          &- \\ 
        DialogueRNN   &39.16     &\textbf{81.69}        &59.77        &67.36       &72.91       &60.27       &64.58          &57.11 \\ 
        DialogueGCN   &\textbf{47.1}      &80.88        &58.71        &66.08       &70.97       &61.21       &65.04          &58.23 \\ \hline
        MMGCN         &42.34     &78.67        &\textbf{61.73}        &\textbf{69.00}       &\textbf{74.33}       &\textbf{62.32}       &\textbf{66.22} &\textbf{58.65}       \\ \hline
    \end{tabular} }
        \caption{ERC performance (F1-score) of different approaches on both IEMOCAP and MELD datasets under the multimodal setting, which means the input includes all the acoustic, visual, and textual modalities; bold font denotes the best performance. Average(w) means weighted average. (The result of CMN and ICON are deficient for suiting two-way conversations only)}
    \label{tab:sota}
\end{table*}

\subsection{Utterance-level Raw Feature Extraction}
The textual raw features are extracted using TextCNN following \cite{hazarika2018icon}. The acoustic raw features are extracted using the \emph{OpenSmile} toolkit with IS10 configuration \cite{schuller2011is10}. The visual facial expression features are extracted using a DenseNet \cite{densenet} pre-traind on the Facial Expression Recognition Plus (FER+) corpus \cite{FER}. 

\subsection{Implementation Details}
The hyperparameters are set as follows: the number of GCN layers are both 4 for IEMOCAP and MELD. The dropout is 0.4. The learning rate is 0.0003. The L2 regularization parameter is 0.00003. $\alpha$, $\eta$ and $\gamma$ are set as 0.1, 0.5 and 0.7 respectively. Considering the class-imbalance in MELD, we use focal loss when training MMGCN on MELD. In addition, we add layer normalization after the speaker embedding.

\subsection{Evaluation Metrics and Significance Test}
Following previous works \cite{hazarika2018icon,majumder2019dialoguernn,ghosal2019dialoguegcn}, we use weighted average f1-score as the evaluation metric. Paired t-test is performed to test the significance of performance improvement with a default significance level of 0.05.

\subsection{Compared Baselines}
\label{sec:baselines}
In order to verify the effectiveness of our model, We implement and compare the following models on emotion recognition in conversation.

\vspace{2pt}
\noindent\textbf{BC-LSTM \cite{poria2017context}:} it
encodes contextual information through Bi-directional LSTM \cite{hochreiter1997long} network. The context-aware features are then used for emotion classification. BC-LSTM ignores speaker information as it doesn't attach any speaker-related information to their model.

\vspace{2pt}
\noindent\textbf{CMN \cite{hazarika2018conversational}:} it
leverages speaker-dependent GRUs to model utterance context combining dialogue history information. The utterance features with contextual information are subject to two distinct memory networks for both speakers. Due to the fixed number of Memory network blocks, CMN can only serve in dyadic conversation scenarios.

\vspace{2pt}
\noindent\textbf{ICON \cite{hazarika2018icon}:}
it extends CMN to model distinct speakers respectively. Same with CMN, two speaker-dependent GRUs are leveraged. Besides, A global GRU is used to track the change of emotion status in the entire conversation and multi-layer memory networks are leveraged to model the global emotion status. Though ICON improves the result of ERC, it still cannot adapt to a multi-speaker scenario.

\vspace{2pt}
\noindent\textbf{DialogueRNN \cite{majumder2019dialoguernn}: }
it models speakers and sequential information in dialogues through three different GRUs, which include Global GRU, Speaker GRU and Emotion GRU. 
Specifically, Global GRU models context information, while Speaker dependent GRU models the status of the certain speaker. The two modules update interactively. Emotion GRU detects emotion of utterances in conversation. Furthermore, in the multimodal setting, the concatenation of acoustical, visual, and textual features is used when the speaker talks, but only use visual features otherwise. However, DialogueRNN doesn't improve much in multimodal settings.

\vspace{2pt}
\noindent\textbf{DialogueGCN \cite{ghosal2019dialoguegcn}:}
it applies GCN to ERC, in which the generated features can integrate rich information. Specifically, utterance-level features encoded by bi-lstm are used to initialize the nodes of the graph, edges are constructed within a certain window. Utterances in the same dialogue but with long distance can be connected directly. Relation GCN\cite{schlichtkrull2018modeling} and GNN\cite{morris2019weisfeiler}, which are both non-spectral domain GCN models, are leveraged to encode the graph. However, DialogueGCN only focuses on the textual modality. 
In order to compare with our MMGCN under the multimodal setting, we extend DialogueGCN by simply concatenating features of three modalities.


\begin{figure*}  
    \centering    
    
    \subfloat[early fusion] 
    {
        \begin{minipage}[t]{0.33\textwidth}
            \centering          
            \includegraphics[width=1\textwidth]{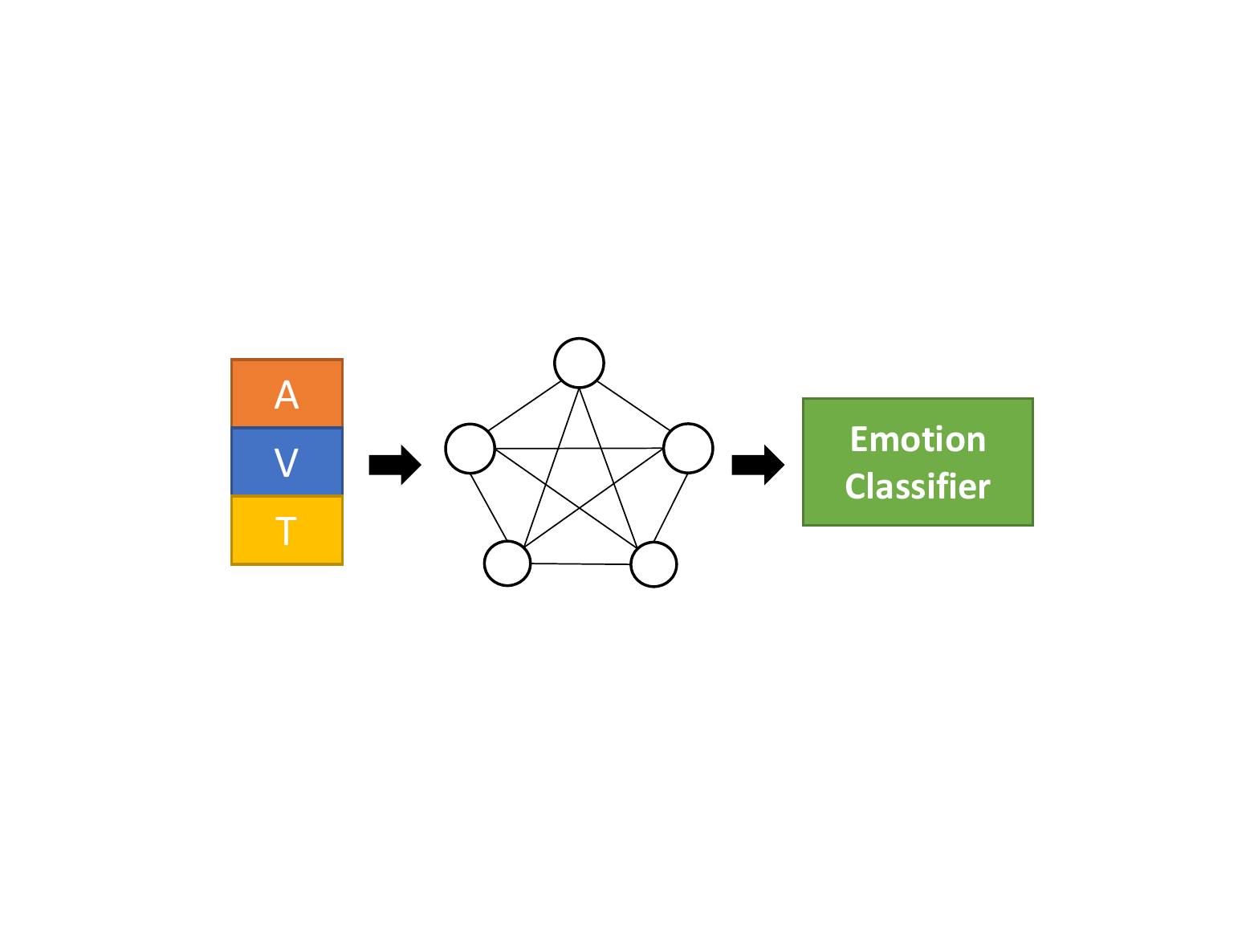}   
        \end{minipage}%
    }
    \subfloat[late fusion] 
    {
        \begin{minipage}[t]{0.33\textwidth}
            \centering      
            \includegraphics[width=1\textwidth]{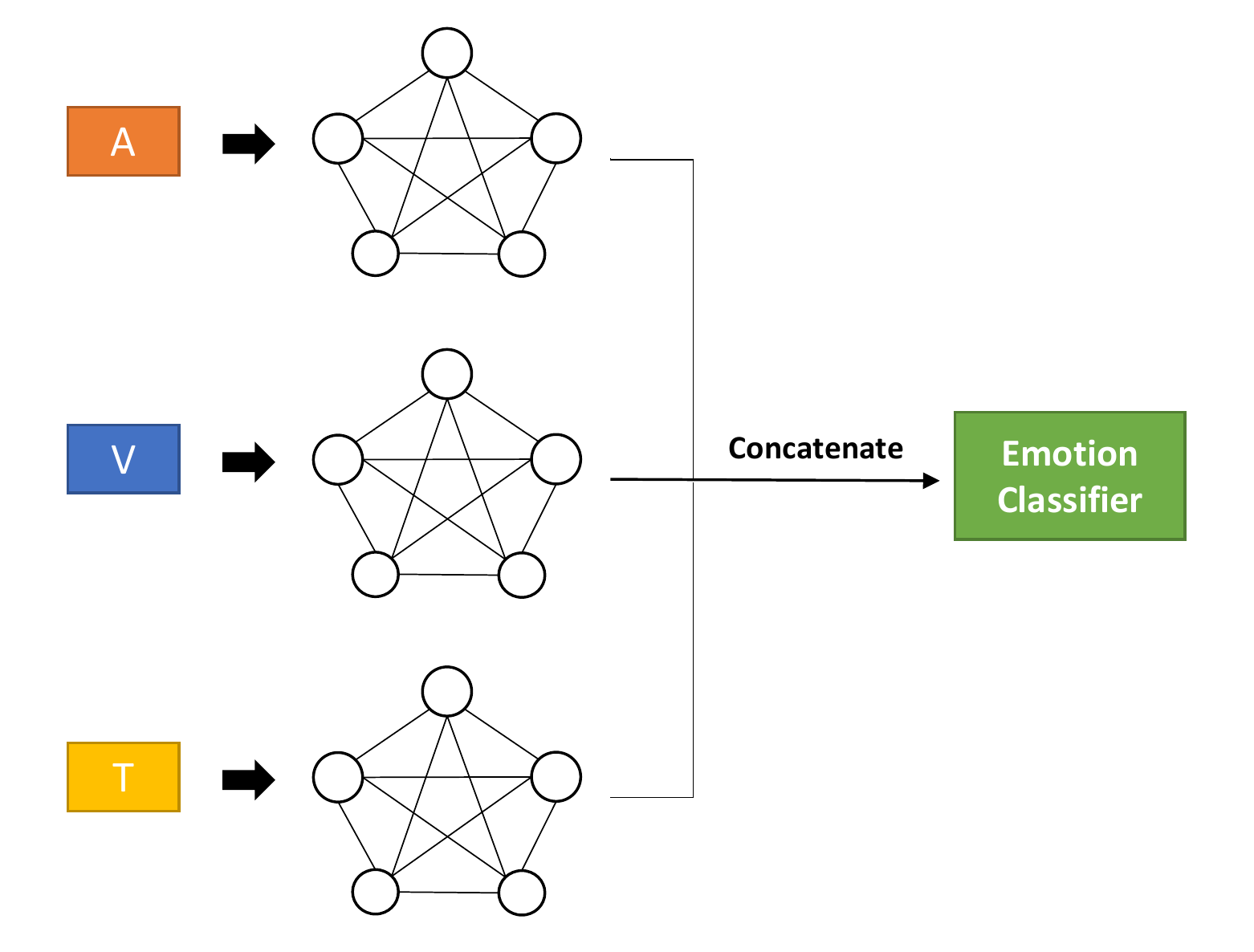}   
        \end{minipage}
    }%
    \subfloat[fusion through gated attention] 
    {
        \begin{minipage}[t]{0.33\textwidth}
            \centering      
            \includegraphics[width=1\textwidth]{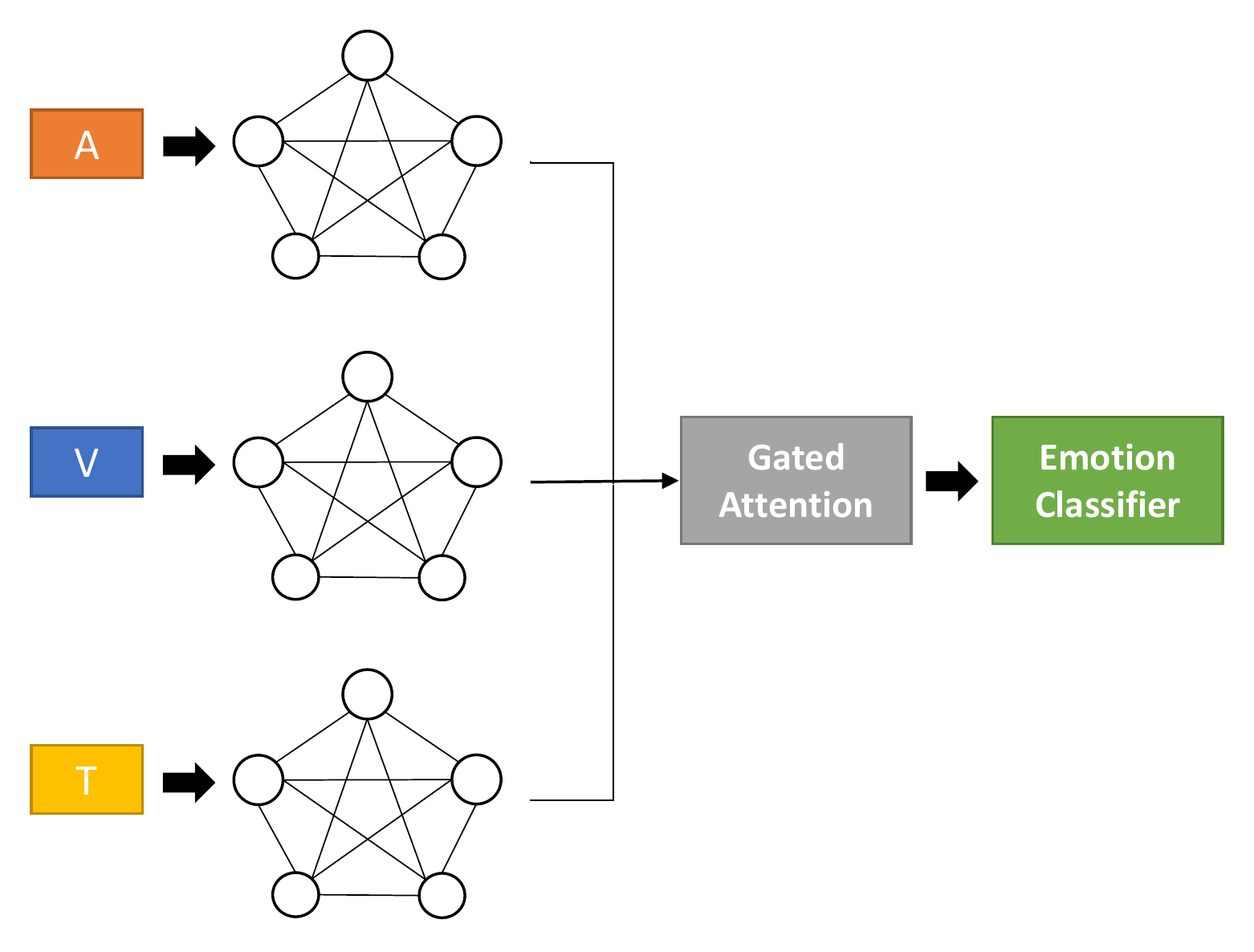}   
        \end{minipage}
    }%

    \caption{Illustration of the three types of multi-modal fusion methods} 
    \label{fig:fusionmethods}  
\end{figure*}

\section{Results and Discussions}
We compare our proposed MMGCN with all the baseline models presented in section~\ref{sec:baselines} on IEMOCAP and MELD datasets under the multimodal setting. In order to compare the results under the same experiment settings, we reimplement the models in the following experiments. 


\subsection{Comparison with other models}
Table~\ref{tab:sota} shows the performance comparison of MMGCN with other models on the two benchmark datasets under the multimodal setting.  
DialougeGCN was the best performing model when using only the textual modality. Under the multimodal setting, DialogueGCN which is fed with the concatenation of acoustic, visual and textual features achieves some slight improvement over the single textual modality. 
Our proposed MMGCN improves the F1-score performance over DialogueGCN under the multimodal setting by absolute 1.18\% on IEMOCAP and 0.42\% on MELD on average, and the improvement is significant with p-value $< 0.05$. 

\subsection{MMGCN under various modality setting}

Table~\ref{tab:modality} shows the performance comparison of MMGCN under different multimodal settings on both benchmark datasets. From Table~\ref{tab:modality} we can see that the best single modality performance is achieved on the textual modality and the worst is on the visual modality, which is consistent with previously reported findings. Adding acoustic and visual modalities can bring additional performance improvement over the textual modality. 
\begin{center}
\begin{table}[t]
    \centering
    \scalebox{0.9}{
    \begin{tabular}{c|cc}
        \hline
        {modality}& {IEMOCAP}& {MELD}             \\ \hline
        a            &54.66          &42.63           \\
        v            &33.86          &33.27           \\
        t            &62.35     &57.72           \\
        at           &65.70     &58.02           \\ 
        vt           &62.89     &57.92           \\ 
        avt          &\textbf{66.22}     &\textbf{58.65}           \\  \hline
    \end{tabular}}
    \caption{ERC performance of MMGCN under different multimodal settings, which means the input contains different combination of the three modalities}
    \label{tab:modality}
\end{table}
\end{center}

\begin{figure*}[t]
\centering
\includegraphics[scale=0.4]{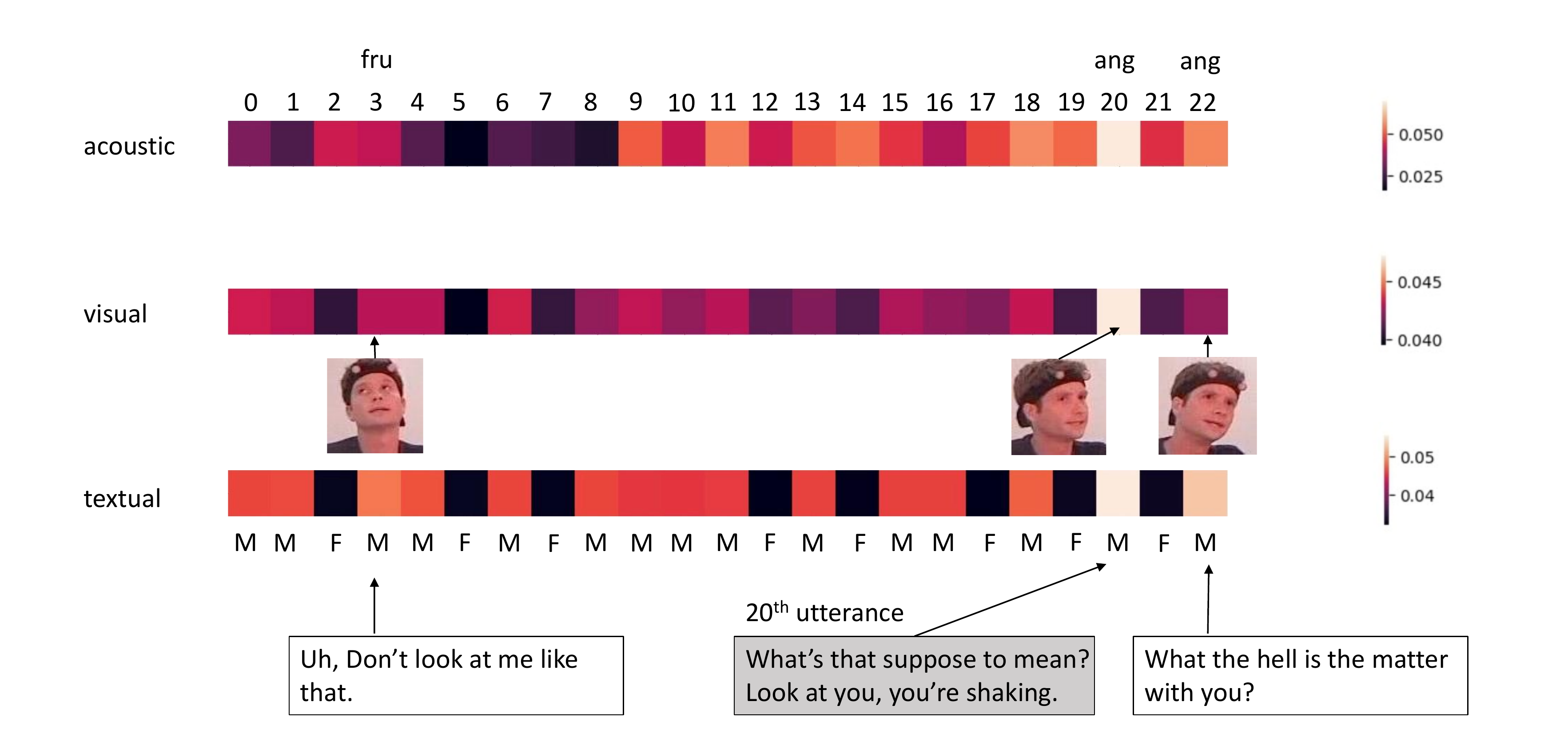}
\caption{Visualization of the heatmap of the adjacent matrix for the $20^{th}$ utterance in a conversation with three modalities. 'M' and 'F' refer to the male and female speakers respectively}
\label{fig:visualize}
\end{figure*}
\subsection{Comparison with other fusion methods}

To verify the effectiveness of MMGCN in multi-modal fusion, we compare it with other multimodal fusion methods, including early fusion, late fusion, fusion through gated attention and other representative fusion methods such as MFN\cite{zadeh2018memory} and MulT\cite{tsai2019multimodal}. The first three fusion methods are illustrated in Figure~\ref{fig:fusionmethods}. 
As for early fusion, multimodal features are concatenated and fed into GCN directly. As for late fusion, features of different modalities are fed into different GCNs respectively and concatenated afterwards. As for fusion through gated attention, features are fed into different GCNs the same way as in late fusion, and then to a gated attention module.
Specifically, the gated attention module can be formulated as follows:

\begin{small}
\begin{align}
\centering
    &r_{i}^{m_j} = tanh(W_{m_j}\cdot h_{i}^{m_j})\\
    &r_{i}^{m_k} = tanh(W_{m_k}\cdot h_{i}^{m_k})\\
    &z = \sigma(W_{z}\cdot h_{i}^{m_j})\\
    &r_{i}^{(m_j, m_k)} = z*r_{i}^{m_j}+(1-z)*r_{i}^{m_k}\\
    &e_{i} = [r_{i}^{(a,v)}, r_{i}^{(a,t)}, r_{i}^{(v,t)}]
\end{align}
\end{small}
where $m_{j}$ and $m_{k}$ could be any modality among \emph{\{a, v, t\}}, $h_{i}^{m_j}$ and $h_{i}^{m_k}$ represent the feature encoded by the corresponding modality encoder, $e_{i}$ represents the final feature representation for the $i^{th}$ utterance.
Considering MFN and MulT are leveraged to fuse multimodal information sequentially, they are used to replace the Modality Encoder. The fused multimodal features are fed to the GCN module subsequently.

Table~\ref{tab:fusion} shows that MMGCN with the graph-based multimodal fusion outperforms all other compared multimodal fusion methods.

\begin{center}
\begin{table}[t]
    \centering
    \scalebox{0.9}{
    \begin{tabular}{c|cc}
        \hline
        & {IEMOCAP}& {MELD}             \\ \hline
        $DeepGCN_{early\_fusion}$           &64.46     &57.94           \\ 
        $DeepGCN_{late\_fusion}$           &64.62     &58.26           \\ 
        $DeepGCN_{gated\_attention}$           &64.45     &58.18           \\ 
        $DeepGCN_{MFN}$                        &62.77           &58.21                   \\
        $DeepGCN_{MulT}$                        &62.37           &57.93                   \\
        $MMGCN$           &\textbf{66.22}     &\textbf{58.65}           \\  \hline
    \end{tabular}}
    \caption{ERC performance comparison of MMGCN and other multimodal fusion methods}
    \label{tab:fusion}
\end{table}
\end{center}

\subsection{MMGCN with different layers}

We investigate the impact of the number of layers in MMGCN on the ERC performance in Table~\ref{tab:layers}. The experiment results show that a different number of layers does affect the ERC recognition performance. Specifically, MMGCN achieves the best performance with 4 layers on both IEMOCAP and MELD.

\begin{center}
\begin{table}[t]
    \centering
    \scalebox{0.9}{
    \begin{tabular}{c|cc}
        \hline
        {layers}& {IEMOCAP}& {MELD}             \\ \hline
        1           &66.12     &58.40           \\ 
        2           &66.17     &58.38           \\ 
        4           &\textbf{66.22}     &\textbf{58.65}           \\ 
        8           &66.10     &58.54           \\ 
        16          &66.06    &58.38           \\
        32          &66.10     &58.42           \\ \hline
    \end{tabular}}
    \caption{ERC performance comparison of MMGCN with different number of layers}
    \label{tab:layers}
\end{table}
\end{center}

\begin{center}
\begin{table}[t]
    \centering
    \scalebox{0.9}{
    \begin{tabular}{c|p{3.5em}<{\centering}p{3em}<{\centering}}
        \hline
        MMGCN & {IEMOCAP}& {MELD}             \\ \hline
        w/ spkr embedding      &\textbf{66.22}   &\textbf{58.65}           \\ 
        w/o spkr embedding  &65.76     &58.38  \\  \hline
    \end{tabular}}
    \caption{Ablation study of the speaker embedding impact on ERC performance}
    \label{tab:speaker}
\end{table}
\end{center}

\subsection{Impact of Speaker Embedding}
~\\
Speaker Embedding can differentiate input features from different speakers. Previous works have reported that speaker information can help improve emotion recognition performance. 
We conduct the ablation study to verify the contribution of speaker embedding in MMGCN as shown in Table~\ref{tab:speaker}. As expected, dropping speaker embedding in MMGCN leads to performance degradation, which is significant by t-test with $p \textless 0.05$. 

\subsection{Case Study}
Fig~\ref{fig:visualize} depicts a scene in which a man and a woman quarrel with each other over a female friend of the man who came to meet with him across 700 miles. They are frustrated or angry in most cases. At the beginning of the conversation, their emotion states are both neutral. Over time, they become emotional. They are both angry at the end of the conversation.
The heatmaps of the adjacent matrix for the $20^{th}$ utterance in the conversation from the three modalities demonstrate that different from simple sequential models, MMGCN pays attention not only to the close context, but also relate to the context in long-distance. For example, as shown in the textual heatmap, MMGCN can successfully aggregate information from the most relevant utterances, even from long-distance utterances, for example the $3^{rd}$ utterance. 
%

\section{Conclusion}
In this paper, we propose an multimodal fused graph convolutional network (MMGCN) for multimodal emotion recognition in conversation (ERC). 
MMGCN provides a more effective way of utilizing both multimodal and long-distance contextual information. It constructs a graph that captures not only intra-speaker context dependency but also inter-modality dependency. With the residual connection, MMGCN can have deep layers to further improve recognition performance. 
We carry out experiments on two public benchmark datasets, IEMOCAP and MELD, and the experiment results prove the effectiveness of MMGCN, which outperforms other state-of-the-art methods by a significant margin under the multimodal conversation setting.


\section{Acknowledgement}
This work was supported by the National Key R$\&$D Program of China under Grant No. 2020AAA0108600, National Natural Science Foundation of China (No. 62072462), National Natural Science Foundation of China (No. 61772535), and Beijing Natural Science Foundation (No. 4192028).

\bibliography{acl2021}
\bibliographystyle{acl_natbib}

\end{document}